# Bringing Image Structure to Video via Frame-Clip Consistency of Object Tokens

**Elad Ben-Avraham**
Tel Aviv University
eladba4@gmail.com

**Roei Herzig**
Tel Aviv University, IBM Research
roeiherz@gmail.com

**Karttikeya Mangalam**
UC Berkeley
mangalam@cs.berkeley.edu

**Amir Bar**
Tel Aviv University
amirb4r@gmail.com

**Anna Rohrbach**
UC Berkeley
anna.rohrbach@berkeley.edu

**Leonid Karlinsky**
MIT-IBM Lab
leonidka@ibm.com

**Trevor Darrell**
UC Berkeley
trevordarrell@berkeley.edu

**Amir Globerson**
Tel Aviv University, Google Research
gamir@tauex.tau.ac.il

## Abstract

Recent action recognition models have achieved impressive results by integrating objects, their locations and interactions. However, obtaining dense structured annotations for each frame is tedious and time-consuming, making these methods expensive to train and less scalable. On the other hand, one does often have access to a small set of annotated images, either within or outside the domain of interest. Here we ask how such images can be leveraged for downstream video understanding tasks. We propose a learning framework StructureViT (SViT for short), which demonstrates how utilizing the structure of a small number of images only available during training can improve a video model. SViT relies on two key insights. First, as both images and videos contain structured information, we enrich a transformer model with a set of *object tokens* that can be used across images and videos. Second, the scene representations of individual frames in video should "align" with those of still images. This is achieved via a *Frame-Clip Consistency* loss, which ensures the flow of structured information between images and videos. We explore a particular instantiation of scene structure, namely a *Hand-Object Graph*, consisting of hands and objects with their locations as nodes, and physical relations of contact/no-contact as edges. SViT shows strong performance improvements on multiple video understanding tasks and datasets, including the first place in the Ego4D CVPR'22 Point of No Return Temporal Localization Challenge. For code and pretrained models, visit the project page at https://eladb3.github.io/SViT/.

## 1 Introduction

Semantic understanding of videos is a key challenge for machine vision and artificial intelligence. It is intuitive that video models should benefit from incorporating scene structure, e.g., the objects that appear in a video, their attributes, and the way they interact. Indeed, several works [5, 25, 26, 34, 35, 42, 80, 82, 88] have demonstrated that incorporating structured representations into models improves both performance and sample efficiency. In particular, [34, 80, 88] showed how to incorporate structured representations by utilizing static image object annotations in videos.

Recently, vision transformers (ViT) [20] have emerged as the leading model for many vision applications [4, 21, 12]. This raises the question: how can structured representations be leveraged for video tasks in a video transformer? Past works [4, 21] have proposed video transformer models but these approaches were not focused on learning structured representations. A recent line of works [34, 80, 88] utilized objects for structured approaches. [82] developed an object-centric model based on region crop features from SlowFast [22], [34] proposed a new block that incorporates object representations, and [88] proposed a structured model for action detection. However, these approaches are limited because they use external pretrained detectors for the frames, which require additional processing time (also during inference). Moreover, they ignore the potential benefits of the connection between the detection and the other tasks, and it is unclear how to use them with more detailed structured annotations. Our approach, described next, builds on these prior works and proposes an effective way to *leverage scene structure in video transformers*, using a relatively *small* set of annotated static images.

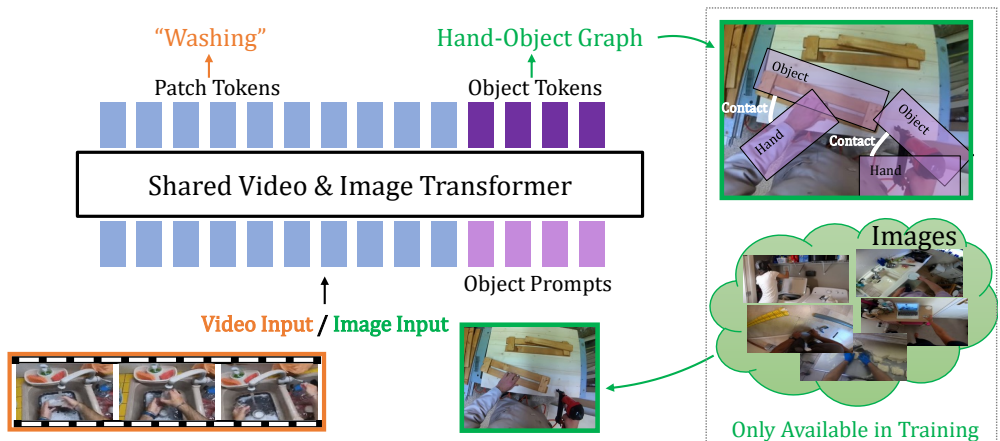

Figure 1: Our SViT approach brings structured scene representations from still images into videos. We use the **HA**nd-**O**bject **G**raph (HAOG) annotations of still images that are only available during training, and videos, which may come from a different domain, with their downstream task annotations. We design a shared video & image transformer that can handle both video and image inputs. During training, given an image, the patch and object tokens are processed together, and the object prompts learn to predict the HAOG, whereas given a video input, the transformer predicts the video label (e.g. *Washing*) based on the patch tokens. During inference, the learned object prompts, which have captured the structured scene information from images, are used to predict the video label.

Most structured transformer-based models for videos require structured annotations during training. For example, frame-level object bounding boxes are considered available in [5, 7, 34, 35, 59, 80]. When it comes to structured annotations of static images, there are numerous datasets with annotations such as boxes, visual relationships, attributes, and segmentations [10, 49, 59]. Since we cannot always expect to have annotated images that perfectly align with our downstream video task, how can these resources be used to build better transformer-based video models? In this work, we propose such an approach that is specifically designed for transformer architectures and offers an effective mechanism for improving video transformers by leveraging images with structured annotations that are *not necessarily associated* with the video domain and can be within or outside the domain of interest.

A natural first step towards image-video knowledge sharing is to use the same transformer model to process both: namely, to view images as "single frame videos", and feed them to a video transformer model. However, this still leaves two key questions: how to model structured information, and how to account for domain mismatch between images and videos. Towards this end, we introduce two key concepts: (i) A set of transformer "object-tokens" – additional tokens initialized from learned embeddings (we refer to these embeddings as "object prompts") that are meant to capture object-centric information in both still images and video. These tokens can be supervised based on the scene structure of images, and that information will also be propagated to videos. To formalize the image structure, we propose the HAnd-Object Graph (HAOG), a simple representation of the interactions between hands and objects in the scene. (ii) A novel "Frame-Clip Consistency loss" that ensures consistency between the "object-tokens" when they are part of a video vs. a still image. Moreover, as we show in our experiments, our approach does not require any alignment between object categories appearing in still images and the ones appearing in the videos of the target downstream task. We name our proposed approach StructureViT (*Bringing Scene **Structure** from Images to **V**ideo via Frame-Clip Consistency of Object **T**okens*, or SViT for short). See Figure 1 for an overview.

Prior work has explored several other video & image models. The I3D model [13] was one of the first successful attempts to leverage ImageNet architectures in video. They proposed to simply convert image 2D classification models into 3D ConvNets by inflating all the filters and pooling kernels.

Here, our focus is on learning the shared structured representations between the two domains in order to transfer this knowledge into videos. Several other methods have explored simultaneous image and video training in the context of multi-task [6] and multi-modal [27] learning. However, these works did not utilize structural information, nor did they promote consistency between image and video predictions – which, as we show, leads to significant gains in downstream video task performance.

To summarize, our main contributions are as follows: (i) we propose a new method for exploiting structured information present in images in order to improve performance of video understanding tasks; (ii) we propose a novel concept of "object tokens" to capture object-centric structure in transformer models of images and videos - a form of learned prompts to a video & image transformer, which during training are associated to the available structure representation of the still images. (iii) we introduce a new Frame-Clip Consistency loss, which promotes consistency between image-level and video-level predictions made by a shared transformer backbone. We show how this loss helps to drive performance improvements in downstream video tasks even in cases when image labels are unrelated to the video task; (iv) we demonstrate improved performance on several video understanding benchmarks, highlighting the effectiveness of the proposed approach.

## 2 The SViT Approach

Our SViT approach learns a structured representation that is exhibited in both static images and video. We consider the setting where the main goal is to learn video understanding downstream tasks while leveraging structured image data. In this paper, we focus on the following video tasks: action recognition, spatio-temporal action detection and object state classification & localization. In training, we assume that we have access to task-specific video labels as well as structured scene annotations for images. Based on the structured representations obtained by using the *object tokens* and regularized by the *Frame-Clip Consistency loss*, the inference is performed only for the downstream video task without explicitly using the structured image annotations.

We begin by describing an image annotation structure we refer to as Hand-Object Graphs (HAOG) (Section 2.1). We then introduce our SViT model (Section 2.2), and the Frame-Clip Consistency loss function (Section 2.3). Our method is schematically illustrated in Figure 2.

### 2.1 The Hand-Object Graph

As mentioned above, our motivation is to bring scene structure from still images into video. In order to achieve this, one component of this work is the construction of a semantic representation of the interactions between the hands and objects in the scene. Specifically, we propose to use a graph-structure we call *Hand-Object Graph* (HAOG), see Figure 1. The nodes of the HAOG represent two hands and two objects with their locations, whereas the edges correspond to physical properties such as contact/no contact. Formally, an HAOG is a tuple $(O, E)$ defined as follows:

**Nodes O:** A set $O$ of $n$ objects. Human-object interaction scenes, which we focus on in this work, usually consist of up to two hands and objects. Thus our node categories are "hand" and "object", and they may also contain attributes (e.g., spatial coordinates). We assume that each image comes with the bounding boxes describing the locations of the two hands, and the objects they interact with. The bounding boxes for the left and right hand are denoted as $b_1, b_2 \in \mathbb{R}^4$, respectively, while for the two corresponding objects interacting with the hands are denoted as $b_3, b_4$. We also assume access to four binary variables $e_1, \ldots, e_4 \in \{0, 1\}$ that indicate whether the corresponding hands or objects in fact exist in the image (e.g., if $e_1 = 0$ the left hand is not in the image, and $b_1$ should be ignored).

**Edges E:** The edges are represented as labeled edges between hand and object nodes. The edge labels describe the physical properties of the scene. In our current modeling, each edge is characterized by a physical property of "contact" or "no-contact". We denote two binary variables $c_1, c_2 \in \{0, 1\}$ that specify if each of the hands is in direct contact with the corresponding object (the left hand $b_1$ can only interact with the object $b_3$, similar for the right hand). In principle, we can also include properties such as distances between objects, 3D hand poses, etc., where available.

We note that images annotated consistent with our defined HAOG format can be found in several datasets [31, 59, 72], and we leverage these in our experiments.

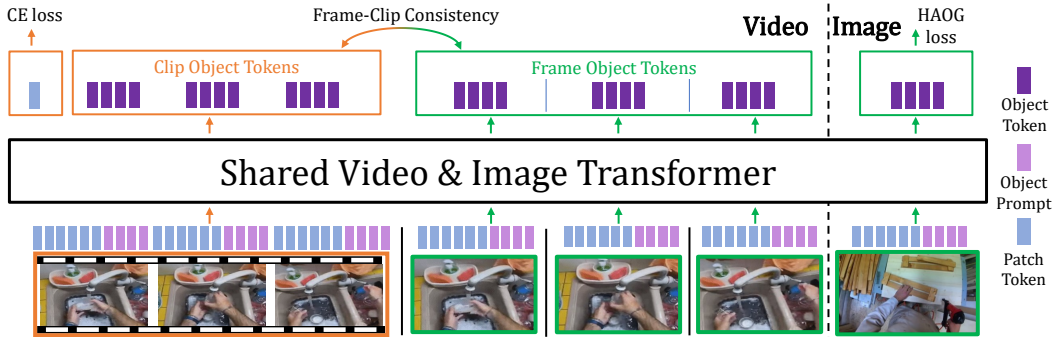

Figure 2: Our shared video & image transformer model processes two different types of tokens: standard patch tokens from the images and videos (**blue**) and the object prompts (**purple**), that are transformed into object tokens (**purple**) in the output. During training, the object tokens (**purple**) are trained to predict the HAOG for still images. For video frames that have no HAOG annotation, we use our "Frame-Clip" loss to ensure consistency between the "frame object tokens" (resulting from processing the frames separately) and the "clip object tokens" (resulting from processing the frame as part of the video). Last, the final video downstream task prediction results from applying a video downstream task head on the average of the patch tokens in the transformer output (after they have interacted with the clip object tokens (**purple**)).

## 2.2 The SViT Transformer Model

We next present the SViT model: a video transformer that employs *object tokens* as a means to capture structure in both images and videos. We begin by reviewing the video transformer architecture, which our model extends. Next, we explain how we process both images and videos. Finally, we describe object tokens and how they are used in videos and still images.

**Shared Video & Image Transformer**. Video transformers [4, 11, 21, 34] extend the Vision Transformer model to the temporal domain. Similar to vision transformers, the input is first "patchified" but with temporally extended 3-D convolution (instead of 2-D) producing a down-sampled tensor $X$ of size $T \times H \times W \times d$. We refer to this as "patch tokens". Then, spatio-temporal position embeddings are added for providing location and time information. Next, multiple stacked self-attention blocks are applied repeatedly on the down-sampled tensor $X$ to produce the final vector representation using mean pooling over the patch tokens.

In our approach, we want to be able to process batches of images or videos. A key desideratum in this context is to be able to input both videos and still-images into the same model. Towards this end, our first change is to rely only on 2-D convolutions in the first transformer step instead of 3-D. This allows single images to be used as inputs without padding, and does not hurt performance in practice.

**Object Tokens**. Our goal is to learn a shared structured representation between the video and the image domains. We do this by adding transformer tokens to represent objects. These tokens are functionally similar to the patch tokens, with two exceptions. First, their initial embedding is learned. We refer to these embeddings as "object prompts". Second, they are used to predict the structured representations when those are available (here, we use the Hand-Object Graphs structure for still images). An object token is used to predict properties of the corresponding node in the HAOG, and the concatenation of two object tokens is used to predict edge properties.

Next, we describe the model more formally. Let $n$ be the maximum number of modeled objects per frame (or still image). We have $n$ tokens for each frame, and thus a total of $T \times n$ tokens. The token for object $i$ at frame $t$ is initialized with the vector $\boldsymbol{o}_i + \boldsymbol{r}_t$ where $\boldsymbol{o}_i \in \mathbb{R}^d$ is a learned object prompt and $\boldsymbol{r}_t \in \mathbb{R}^d$ is a learned temporal positional embedding (the same one used for initializing the patch tokens). With these new tokens, we have $T \times (H \times W + n)$ tokens (i.e, vectors in $\mathbb{R}^d$), and these together will go through the standard self-attention layers.

We denote the operator that outputs the final representation of the object tokens by $F_O$, where for a video $V$ we have $F_O(V) \in \mathbb{R}^{T \times n \times d}$ and for an image $I$ we have $F_O(I) \in \mathbb{R}^{n \times d}$. The $F_O(I)$ representation is used to predict structured representations for single images. We also include a loss that makes single frame representations align with those of clip representations (see Section 2.3). The operator that outputs the final representation vector used for the video downstream task is denoted by $F_{CLS}$, where for video $V$ we have $F_{CLS}(V) \in \mathbb{R}^d$.

Finally, our method can be used on top of the most common video transformers (MViT [21], TimeSformer [11], Mformer [66], Video Swin [56]). For our experiments, we use the MViTv2 [53] model because it performs well empirically.

## 2.3 Losses and Training

During training we have a set of images annotated with HAOGs and videos with downstream task labels. We use these as inputs to our model (Section 2.2) and optimize the losses described below.

**Video loss**. Each training video $V$ has a corresponding ground-truth category $Y \in \{1, \ldots, K\}$. As noted above, we have a vector representation for the entire video denoted by $F_{CLS}(V)$. Thus, we simply use a neural network to predict logits for $Y$. Specifically, our predicted logits are: $\hat{Y} = FC(F_{CLS}(V)) \in \mathbb{R}^K$, where $FC$ is a trainable fully connected network. We then consider a standard cross-entropy between the predicted logits $\hat{Y}$ and true labels $Y$.

$$\mathcal{L}_{Vid} := \text{CE}(\text{Softmax}(Y, \hat{Y})) , \quad (1)$$

**HAOG loss**. The transformer outputs a set of $n$ object tokens for each image. Since we have supervision for these objects, we predict structured representations from the tokens, and introduce a loss to optimize these predictions. Recall our notation from Section 2.1. We let the transformer use $n = 4$ object tokens which correspond to the two hands (bounding boxes $\boldsymbol{b}_1, \boldsymbol{b}_2$) and two manipulated objects (bounding boxes $\boldsymbol{b}_3, \boldsymbol{b}_4$). The $j^{th}$ token is used to predict a corresponding bounding box $b_j$, as well as the existence variable $e_j$. Formally, we let:

$$\hat{\boldsymbol{b}}_j = \text{FC}_{bb}(F_O^j(I)) , \ \hat{e}_j = \text{FC}_e(F_O^j(I)) , \quad (2)$$

where $FC_{bb}, FC_e$ are fully connected networks with four and one output respectively, $F_O^j(I)$ is the $j^{th}$ object token, and $I$ is the input image. Formally, this defines the loss of the nodes as follows:

$$\mathcal{L}_{Nodes} = \sum_{i=1}^{4} \text{BCE}(\text{Sigmoid}(\hat{e}_i), e_i) + e_i \left( \text{GIoU}(\hat{\boldsymbol{b}}_i, \boldsymbol{b}_i) + L_1(\hat{\boldsymbol{b}}_i, \boldsymbol{b}_i) \right) \quad (3)$$

where L1 is the standard L1 distance, and GIoU is used as in [69]. Furthermore, we predict the contact variable from the concatenation of the corresponding hand and object tokens. Namely for $j \in \{1, 2\}$:

$$\hat{c}_j = \text{Softmax}(\text{FC}_c([F_O^j(I), F_O^{j+2}(I)])) \quad (4)$$

where $FC_c$ is a fully connected with one output. Thus, we define the loss of the edges as follows:

$$\mathcal{L}_{Edges} = \sum_{i=1}^{2} \text{CE}(\hat{c}_i, c_i) \quad (5)$$

The final prediction loss is the sum of the node and edge losses:

$$\mathcal{L}_{HAOG} = \mathcal{L}_{Nodes} + \mathcal{L}_{Edges} \quad (6)$$

**Frame-Clip Consistency Loss**. Since we have different losses for images and video, the model could find a way to minimize the image loss in a way that only applies to the images and does not transfer to video.[1] To avoid this, we need to make sure that the video representation contains the same type of information as the still images, and in particular the structured information carried by the object tokens. To achieve this, we explicitly enforce the object tokens to be consistent across still images (frames) and videos (clips) using a frame-clip consistency loss. To calculate the loss, we process each video twice: once as a clip, and once as a batch of $T$ frames. Namely, a clip $V$ consisting of $T$ frames $I_1, ..., I_T$, will be processed once as $F_O(V)$ and once as a list of images $(F_O(I_1), ..., F_O(I_T))$. This results in two groups of $T \times n$ object tokens, the first are the *clip object tokens*, denoted as, $\mathcal{O}_{clip} := F_O(V) \in \mathbb{R}^{T \times n \times d}$, the second are the *frame object tokens*, denoted as $\mathcal{O}_{frames} := [F_O(v_1), ..., F_O(v_T)] \in \mathbb{R}^{T \times n \times d}$. Since both groups originate from the same input, each clip object token has a single corresponding frame object token. Namely, there is an alignment between the clip object tokens and the frame object tokens. Following the intuition that each clip object token should contain the information from its corresponding frame object token, we minimize the $L1$ distance between each clip object token and frame object token. Namely:

$$\mathcal{L}_{Con} := L_1(\mathcal{O}_{\text{clip}}, \mathcal{O}_{\text{frames}}) \qquad (7)$$

---

[1]For example, it could "overfit" to relying on the positional embedding of the first frame only when minimizing image-related losses.

**Overall loss.** The total loss consists of $\mathcal{L}_{Con}$, $\mathcal{L}_{HAOG}$ and $\mathcal{L}_{Vid}$, where $\mathcal{L}_{Vid}$ is the loss of the main task. Each of the three terms in the loss is multiplied by a hyper-parameter ($\lambda_{Con}$, $\lambda_{HAOG}$, $\lambda_{Vid}$), and the total loss is the weighted combination of the three terms:

$$\mathcal{L}_{\text{Total}} := \lambda_{Con}\mathcal{L}_{Con} + \lambda_{HAOG}\mathcal{L}_{HAOG} + \lambda_{Vid}\mathcal{L}_{Vid} \qquad (8)$$

## 3 Experiments and Results

We begin by describing the datasets, implementation details, and baselines and variants. We then evaluate our approach on several benchmarks. Specifically, we consider the following tasks: Compositional Action Recognition (Section 3.4), Object State Change Classification & Localization (Section 3.5), Action Recognition (Section 3.6), and Spatio-Temporal Action Detection (Section 3.7).

### 3.1 Datasets

We first describe the datasets used for the downstream video tasks, and those used as "auxiliary" datasets with annotated images. We use the following video datasets: **(1) Something-Something v2 (SSv2)** [30] is a dataset containing 174 action categories of common human-object interactions. **(2) SomethingElse [59]** which exploits the compositional structure of SSv2, where a combination of a verb and a noun defines an action. We follow the official compositional split from [59], which assumes the set of noun-verb pairs available for training is disjoint from the set given at test time. **(3) Ego4D** [31] is a new large-scale dataset of more than 3,670 hours of video data, capturing the daily-life scenarios of more than 900 unique individuals from nine different countries around the world. **(4) Diving48** [54] contains 48 fine-grained categories of diving activities. **(5) Atomic Visual Actions (AVA)** [32] is a benchmark for human action detection, we report Mean Average Precision (mAP) on AVA-V2.2. For "auxiliary" image datasets we use the **Ego4D** [31], and the **100 Days of Hands (100DOH)** [72] datasets. We collected 79,921 annotated images from 100DOH, and 57,213 annotated images from Ego4D.[2] The AVA or SSv2 datasets are also used as "auxiliary" images in some experiments. For more details, see Section D.1 in the supplementary. For the image datasets, the image annotations are based on frames selected from videos, but these are often sparsely selected from the videos, so it is natural to treat them as annotated still images.[3]

### 3.2 Implementation Details

SViT is implemented in PyTorch, and the code is available on our project page. Our training recipes and code are based on the MViTv2 model, and were taken from https://github.com/facebookresearch/mvit. We pretrain the SViT model on the K400 [45] *video* dataset with our *auxiliary image* datasets. Then, we finetune on the target video understanding task (detailed in Section 3.1) together with the *auxiliary image* datasets and the SViT loss. Each training batch contains 64 images and 64 videos in order to minimize the overall loss in Equation 8. For more details about datasets and evaluation see Section C of Supplementary.

### 3.3 Baselines and SViT variants

**Baselines.** In the experiments, we compare SViT to several models from prior work reported for the corresponding datasets. These include the following approaches: *I3D* [13], *SlowFast* [22], and the state-of-the-art transformers – *Mformer* [66] and *MViTv2* [53]. In addition, we provide the *MViTv2 multi-task (MViTv2 MT)* baseline for comparing a naive multi-tasking approach to our model. This is a version of MViTv2 which performs video classification and HAOG prediction from images. The MViTv2 model is enhanced with two heads: the first head is composed of an MLP that predicts the video labels given a final representation vector, the second head consists of MLPs that predict the HAOG information (i.e., hands and object boxes, and contact information).

---
[2] Comparatively, SomthingElse contains 2,550,700 frames, showing that image datasets are relatively small.
[3] The SomethingElse/SSv2 datasets has annotations for all clip frames, but we show in the experiments that SViT only needs a small percentage of these.

| Model | Boxes Annotations | Compositional Top-1 | Top-5 | Base Top-1 | Top-5 | Few-Shot 5-Shot | 10-Shot |
|---|---|---|---|---|---|---|---|
| SlowFast [22] | ✗ | 45.2 | 73.4 | 76.1 | 93.4 | 22.4 | 29.2 |
| TimeSformer [11] | ✗ | 44.2 | 76.8 | 79.5 | 95.6 | 24.6 | 33.8 |
| Mformer [66] | ✗ | 60.2 | 85.8 | 82.8 | 96.2 | 28.9 | 33.8 |
| MViTv2 [53] | ✗ | 63.3 | 87.5 | 83.7 | 96.8 | 32.7 | 40.2 |
| MViTv2 MT | ✓ | 63.8 | 87.3 | 85.2 | 97.1 | 33.7 | 41.4 |
| SViT-SFT | ✓ | 64.2 | 87.9 | 85.0 | 97.1 | 33.8 | 41.7 |
| SViT-DD | ✓ | 65.1 | 88.0 | 85.8 | 97.4 | 34.6 | 42.5 |
| SViT ×2% | ✓ | 65.6 | 88.1 | 85.9 | 97.4 | 34.4 | 43.1 |
| SViT-ID | ✓ | 65.8 | 88.3 | 85.6 | 97.3 | 34.4 | 42.6 |

Table 1: **Compositional and Few-Shot Action Recognition** on the "SomethingElse" dataset.

**SViT Variants**. As mentioned above, we use several sources of auxiliary image datasets. In some cases, we have auxiliary images that are frames from video understanding datasets. For example, in Ego4D, we have both videos and frames annotated from some of these videos. In this sense, the auxiliary data is "in-domain" for some of the video tasks. In order to evaluate the importance of this effect, we explore "Different Domain" (*SViT-DD*) training, which avoids using images from a given video task as auxiliary during finetuning and pretraining. Similarly, "In Domain" (*SViT-ID*) refers to the case where we use only images from the video task during finetuning, but not necessarily during pretraining . We consider another variant that does not use the auxiliary images during finetuning at all (but does use them during pre-training), and refer to it as *SViT-SFT*. Finally, to explore the effect of data-size in the "In Domain" setting, we evaluate a version *SViT ×X%* that only uses X% of the in-domain finetuning data. For MViTv2 MT, the auxiliary images correspond to the "In Domain" setting (i.e., same setting as for SViT-ID).

**Auxiliary Images datasets**. For SViT-DD, we generally use 100DOH and Ego4D for pretraining and finetuning, while for the Ego4D tasks, we only use 100DOH. SViT-ID uses 100DOH and Ego4D for pretraining and in-domain downstream frames for finetuning. The SViT-SFT pretraining setting is similar to that of SViT-DD. SViT ×X% setting is similar to that of SViT-ID. For more details, see Section D.1 in the supplementary.

### 3.4 Compositional & Few-Shot Action Recognition

Several video datasets define an action as a combination of a verb and a noun. One challenge then is to recognize combinations that were not seen during training. This "compositional" setting was explored in the "SomethingElse" dataset [59], where verb-noun combinations in the test data do not occur in the training data. The split contains 174 classes with 54,919/54,876 videos for training/validation. We also evaluate on the few-shot compositional action recognition task in [59] (See Section C.3 in supplementary).

Table 1 reports the results for these two tasks. All of our variants outperform all prior works and the MViTv2 MT on the *Compositional* and *Few-shot* task. From the table, we can see that using 2% of the in-domain data as auxiliary images (SViT×2%), is almost comparable to using all the available in-domain images (SViT-ID). Furthermore, the results are comparable to SViT-DD (0.7 difference) when trained on auxiliary images from different domains. This suggests that our approach optimizes the use of structural information across domains. Last, SViT-DD outperforms SViT-SFT when using the same data, demonstrating the value of finetuning video and image together.

### 3.5 Object State Change Classification and Localization

Hands and objects are key elements in human activity. Two tasks related to hand-object interaction have recently been introduced in the Ego4D [31] dataset. The first is temporal localization, which is defined as finding key frames in a video clip that indicate a change in object state. The second is object state change classification, which indicates whether an object state has changed or not. SViT won the first place at the Ego4D CVPR'22 Point of No Return Temporal Localization Challenge.

| (a) Object State Change in Ego4D. | | | (b) Action Recognition in SSv2. | | (c) Action Recognition in Diving48. | |
|---|---|---|---|---|---|---|
| Model | Temporal loc. error | Cls. top-1 | Model | Top-1 | Model | Top-1 |
| Bi-LSTM [†] | 0.790 | 65.3 | SlowFast, R101[†] | 63.1 | SlowFast, R101 [†] | 77.6 |
| BMN[†] | 0.780 | - | ViViT-L [†] | 65.4 | TimeSformer [†] | 74.9 |
| I3D ResNet-50[†] | 0.739 | 68.7 | MViTv1 [†] | 64.7 | TimeSformer-HR [†] | 78.0 |
| MViT-v2[†] | 0.702 | 71.6 | MViTv2 [†] | 68.2 | MViTv2 [†] | 73.1 |
| MViTv2 MT | 0.678 | 71.1 | MViTv2 MT | 68.3 | MViTv2 MT | 75.5 |
| SViT-SFT | 0.654 (−.048) | 70.4 (−1.2) | SViT-SFT | 68.9 (+0.7) | SViT-SFT | 77.9 (+4.8) |
| SViT-DD | 0.646 (−.056) | 73.6 (+2) | SViT-DD | 69.2 (+1) | SViT-DD | 79.8 (+6.7) |
| SViT ×2% | 0.649 (−.053) | 73.4 (+1.8) | SViT ×2% | 69.4 (+1.2) | | |
| SViT-ID | 0.64 (−.062) | 73.8 (+2.2) | SViT-ID | 69.7 (+1.5) | | |

Table 2: **Results on Ego4D, SSv2, and Diving48 datasets.** Evaluation metrics for the Ego4D object state change task are absolute error in seconds (temporal loc. error. Lower is better) for temporal localization, and top-1 accuracy (Cls. top-1) for state change classification. We denote methods that do not use any additional structured annotation with [†].

Table 2a reports results on the temporal localization and object state change classification tasks on the Ego4D dataset. We observe that SViT-ID, SViT×2% and SViT-DD perform better than SViT-SFT and are very close. These results are consistent with those presented in Section 3.4, which indicates that our method successfully leverages structural information across domains, and that a joint image and video finetuning is beneficial.

### 3.6 Action Recognition

Table 2b and Table 2c report the results for standard action recognition task on the SSv2 and Diving48 datasets respectively. It can be seen that in Diving48 dataset, our method improves over the MViT-v2 baseline by 6.7%, outperforming all the methods. This demonstrates that our method has some ability to generalize the learned structural information of hand-object interactions even to fine-grained "human-motion" tasks such as Diving48. We also note that Diving48 has no structured annotations available, thus SViT-ID (and SViT×2%) cannot be provided. On SSv2, SViT-ID improves the baseline by 1.5%, SViT×2% improves by 1.2%, and SViT-DD improves by 1.0%. As we observed significant improvements on the more challenging SomethingElse setup, we hypothesize that SomethingElse is more likely to benefit from the ability to generalize to unseen combinations of objects and hands exhibited by SViT. We provide visualizations of the HAOGs predicted by SViT in Section B of the supplementary.

### 3.7 Spatio-temporal Action Detection

Table 3 reports the results for spatio-temporal action detection on the AVA dataset. In the literature, the action detection task on AVA is formulated as a two stage prediction procedure. The first step is the detection of bounding boxes, which are obtained through an off-the-shelf pretrained person detector. The second step involves predicting the action being performed at each detected bounding box. The performance is benchmarked on the end result of these steps and is measured with the Mean Average Precision (MAP) metric. Typically, for fair comparison, the detected person boxes are kept identical across approaches and hence the final performance depends directly on the ability of the approach to utilize the video and box information. We observe that all SViT variants improve over the MViT-v2 baseline. Specifically, SViT-ID improves the MViT-v2 baseline by 1.7, SViT-DD improves by 1.5, and SViT-SFT improves by 1.0.

| Model | mAP |
|---|---|
| SlowFast [22] [†] | 23.8 |
| MViTv2 [†] | 26.8 |
| MViTv2 MT | 27.2 |
| SViT-SFT | 27.8 (+1.0) |
| SViT-DD | 28.3 (+1.5) |
| SViT-ID | 28.5 (+1.7) |

Table 3: **Spatio-temporal Action Detection** on AVA-V2.2.

### 3.8 Ablations

We perform a comprehensive ablation study on the compositional action recognition task [59] on the SomethingElse dataset to measure the contribution of the different SViT components (Table 4).

| (a) Method components | | (b) Images Amount | | (c) Absence of hands/objects | | | |
|---|---|---|---|---|---|---|---|
| Model | Top-1 | % of data | Top-1 | Model | Objects | Hands | Top-1 |
| MViTv2 | 63.3 | 2 | 65.6 | MViTv2 | ✓ | ✗ | 64.6 |
| MViTv2 MT | 63.8 | 25 | 65.6 | SViT-ID | ✓ | ✗ | 66.5 |
| MViTv2+Object Tokens (OT) | 65.0 | 50 | 65.7 | MViTv2 | ✗ | ✓ | 64.9 |
| MViTv2 +OT+FC Loss (SViT) | 65.8 | 100 | 65.8 | SViT-ID | ✗ | ✓ | 67.0 |

Table 4: **Ablations.** We show (a) Contributions of SViT components. (b) Amount of annotated images used in training. (c) Examine the robustness of SViT to clips that does not contain hands/contains only hands. The experiments are performed on the SomethingElse split. For more ablations, see A in supplementary.

**Components of the SViT model**. In Table 4a, we ablate our model to show the significance of the individual components. First, we train a standard MViTv2 in a multi-task setting (MViTv2 MT), where we use videos for video action recognition, and images for HAOG prediction. There is a small improvement (+0.5) over the baseline. We then take the MViTv2 and add the object tokens and corresponding SViT loss; this leads to a more significant improvement (+1.7). Last, when adding both the consistency loss and the object tokens, we observe the largest improvement (+2.5).

**The amount of in-domain data**. In Table 4b, we demonstrate the effect of the amount of in-domain image data used during training. Specifically, since "SomethingElse" includes HAOG annotations for each frame, we train our method with different amount of images (2%, 25%, 50%, and 100%) from the "SomthingElse" dataset. It can be seen that our approach takes advantage of the structured information within a small amount of data and, as a result, has enhanced sample-efficiency.

**Robustness to absence of hands/objects**. In table Table 4c we verify whether our approach works in the cases where no hands or objects are visible, we conducted experiments on SomethingElse (ground-truth annotations available): (i) After filtering the videos containing (annotated) hands, we tested our model and the MViTv2 model on the filtered split. MViTv2 achieved an accuracy of 64.6 while our model achieved 66.5. This implies an improvement of +1.9. (ii) After filtering the videos containing (annotated) objects in more than 40% of the frames, we tested our model and the MViTv2 model on the filtered split. MViTv2 achieved an accuracy of 64.9 while our model achieved 67.0. This implies an improvement of +2.1. We can observe that our model outperforms the baseline even when there are no objects or hands in the videos, demonstrating the robustness of our approach. The total improvement (+1.9 and +2.1) is also a little bit lower than before the filtering (+2.5), which indicates that there has been a slight degradation. Nevertheless, hand-object graphs remain valuable.

**How important are the HAOGs**. To examine how important is the information provided by the HAOG, we suggest learning HAOGs without any useful information. Thus, we run an experiment in which the HAOGs are completely random. This means that, for each image, a random HAOG is generated by sampling the boxes and their relationships uniformly. We refer to this experiment as SViT-Random-ID. On the SomethingElse dataset, the SViT-Random-ID result is 50.6 while the SViT-ID is 65.8. This shows that predicting actual HAOG attributes provides important information that can be helpful for video downstream tasks.

## 4 Related Work

**Joint Training with Images and Videos**. Vision Transformers [20] are versatile architectures designed to be flexible in terms of input sequence length. Aside from the data "'patchification'" layer, the rest of the transformer is agnostic to the domain of the input. Consequently, an exciting direction of joint training image-video has emerged in computer vision, a direction previously less natural with convolutional architectures. Specifically, this allows for training networks simultaneously on image and video, with most of the network parameters shared across both domains. Recently, multi-modal models became widely popular, from shared backbones [15, 57, 61] to separate encoders for each input modality [74, 77], and even non-vision modalities [39, 40]. In our case, we focus on video and image input, similar to [6, 27, 86]. In contrast to these works, our training approach utilizes two domains, while one of them (image) is used only during training and is supervised by a different task. Additionally, as we demonstrate here, we are not greatly impacted by domain mismatches between images and videos, as in previous works. In a related area of research, multi-task learning [14] involves developing models that predict output for multiple tasks using a common input. Our work is different since it uses two inputs from different domains, images and videos. Similarly, our work can be viewed as a form of auxiliary task, as already shown in many works [29, 38, 73]. However, it differs in that the shared object tokens are utilized to learn the shared structured representation between video and image domains, which enhances the self-attention layers with the structured representation for the main video task.

**Learning from video using other modalities**. Researchers have proposed different approaches to learn from videos using audio and natural language [1, 3, 19, 62, 64, 65, 68, 81]. Vision tasks like Video Captioning [19, 24, 65, 81] and Visual Quesion Answering [2, 78] require understanding of both visual and natural language. Another line of vision and language works explored object labels, supervision of textual descriptions, or the use of region crops [50, 52], in contrast to SViT, which does not require any of these components. Additionally, since high quality captions are expensive, researchers have proposed self-supervised ways to learn from both modalities without labeled data [62, 61, 68]. In these previous works, the two modalities must be consistent and aligned (e.g., text, audio, and RGB should be consistent for same sample). However, in our work, we do not require that the annotated images be precisely aligned with our downstream video task. Last, other new works [16, 70] explored different structured representations for videos, different from what we do, which is to leverage structure from images to videos.

**Structured Models**. Structured models have recently been successfully applied in many computer vision applications: video relation detection [55, 71, 76], vision and language [17, 51, 52, 77], relational reasoning [8, 9, 36, 48, 41, 67, 84, 85], human-object interactions [23, 44, 83], and even image & video generation [7, 33, 43]. A recent line of works [5, 25, 26, 34, 35, 42, 58, 60, 63, 75, 79, 80, 82, 88] focuses on capturing spatio-temporal interactions for video understanding. Some of these works [34, 80, 82, 88] used static image object annotations to build the representations. However, these works differ from ours. [82] proposed the Object Transformers (OT) for long form video understanding, which model a video as a set of objects by extracting region crops from SlowFast [22] based on bounding boxes from an external detector. Contrary to this work, our transformer contains both patch tokens and object tokens, whereas the OT contains only object tokens. Additionally, our model learns the object-centric representation without the need for an external detector. Another work [88], proposed a structured model for action detection that explicitly utilizes image annotations. They use a GCN [47] to reason about actors and objects interaction, on top of an I3D network. Unlike this work, we examine how image annotations can be exploited as part of a video transformer model.

**Video Understanding Models**. We work on several well-benchmarked down-stream video tasks such as compositional action recognition on Something-Else dataset [59] and action recognition with focus on temporal cues de-emphasizing appearance such as in Something-Something V2 dataset [30] and Diving48 [54]. Several recent video transfromer works performed well on SSv2, such as ViViT [4], MViT [21], MFormer [66], TimeSformer [11], MViTv2 [53], and Video Swin [56]. We choose to work with MViTv2, although our method can be used on top of any of these. Additionally, we also adapt our proposed method to interesting new tasks on the Ego4D dataset [31]. We believe such fine-grained video understanding tasks require richer scene and object semantic reasoning, a requirement well afforded by our proposed Hand-Object Graph.

## 5 Conclusion

Video understanding is a key element of human visual perception, but modeling remains a challenge for machine vision. In this work, we demonstrated the importance of learning from the scene structure of a small set of images to facilitate video learning within or outside the domain of interest. According to our empirical study, our SViT approach improves performance on four video understanding tasks. Note, that we did not prioritize making the HAOG annotations richer, i.e., it may be possible to add other physical properties to improve structure modeling. In addition, we noticed that leveraging the structure is effective even with $2\%$ of the in-domain frames, as well as when applying it from outside the domain. We leave for future work the challenge of utilizing different physical properties and more complex structures. Regarding potential impact of the method, We do not anticipate a specific negative impact, but, as with any Machine Learning method, we recommend to exercise caution.

**Acknowledgements**

This project has received funding from the European Research Council (ERC) under the European Unions Horizon 2020 research and innovation programme (grant ERC HOLI 819080). Prof. Darrell's group was supported in part by DoD including DARPA's Semafor, PTG and/or LwLL programs, as well as BAIR's industrial alliance programs.

# References

[1] Triantafyllos Afouras, Andrew Owens, Joon Son Chung, and Andrew Zisserman. Self-supervised learning of audio-visual objects from video. In *European Conference on Computer Vision*, pages 208–224. Springer, 2020.

[2] Stanislaw Antol, Aishwarya Agrawal, Jiasen Lu, Margaret Mitchell, Dhruv Batra, C Lawrence Zitnick, and Devi Parikh. Vqa: Visual question answering. In *Proceedings of the IEEE international conference on computer vision*, pages 2425–2433, 2015.

[3] Relja Arandjelovic and Andrew Zisserman. Look, listen and learn. In *Proceedings of the IEEE International Conference on Computer Vision*, pages 609–617, 2017.

[4] Anurag Arnab, Mostafa Dehghani, Georg Heigold, Chen Sun, Mario Lučić, and Cordelia Schmid. Vivit: A video vision transformer, 2021.

[5] Anurag Arnab, Chen Sun, and Cordelia Schmid. Unified graph structured models for video understanding. In *ICCV*, 2021.

[6] Max Bain, Arsha Nagrani, Gül Varol, and Andrew Zisserman. Frozen in time: A joint video and image encoder for end-to-end retrieval. In *IEEE International Conference on Computer Vision*, 2021.

[7] Amir Bar, Roei Herzig, Xiaolong Wang, Anna Rohrbach, Gal Chechik, Trevor Darrell, and A. Globerson. Compositional video synthesis with action graphs. In *ICML*, 2021.

[8] Fabien Baradel, Natalia Neverova, Christian Wolf, Julien Mille, and Greg Mori. Object level visual reasoning in videos. In *ECCV*, pages 105–121, 2018.

[9] Peter W Battaglia, Jessica B Hamrick, Victor Bapst, Alvaro Sanchez-Gonzalez, Vinicius Zambaldi, Mateusz Malinowski, Andrea Tacchetti, David Raposo, Adam Santoro, Ryan Faulkner, et al. Relational inductive biases, deep learning, and graph networks. *arXiv preprint arXiv:1806.01261*, 2018.

[10] Rodrigo Benenson, Stefan Popov, and Vittorio Ferrari. Large-scale interactive object segmentation with human annotators. In *CVPR*, 2019.

[11] Gedas Bertasius, Heng Wang, and Lorenzo Torresani. Is space-time attention all you need for video understanding? In *Proceedings of the International Conference on Machine Learning (ICML)*, July 2021.

[12] Nicolas Carion, Francisco Massa, Gabriel Synnaeve, Nicolas Usunier, Alexander Kirillov, and Sergey Zagoruyko. End-to-end object detection with transformers. In Andrea Vedaldi, Horst Bischof, Thomas Brox, and Jan-Michael Frahm, editors, *Computer Vision – ECCV 2020*, pages 213–229, 2020.

[13] Joao Carreira and Andrew Zisserman. Quo vadis, action recognition? a new model and the kinetics dataset. In *proceedings of the IEEE Conference on Computer Vision and Pattern Recognition*, pages 6299–6308, 2017.

[14] Rich Caruana. Multitask learning. In *Encyclopedia of Machine Learning and Data Mining*, 1998.

[15] Lluís Castrejón, Yusuf Aytar, Carl Vondrick, Hamed Pirsiavash, and Antonio Torralba. Learning aligned cross-modal representations from weakly aligned data. *2016 IEEE Conference on Computer Vision and Pattern Recognition (CVPR)*, pages 2940–2949, 2016.

[16] Brian Chen, Xudong Lin, Christopher Thomas, Manling Li, Shoya Yoshida, Lovish Chum, Heng Ji, and Shih-Fu Chang. Joint multimedia event extraction from video and article. In *EMNLP*, 2021.

[17] Yen-Chun Chen, Linjie Li, Licheng Yu, Ahmed El Kholy, Faisal Ahmed, Zhe Gan, Yu Cheng, and Jingjing Liu. Uniter: Universal image-text representation learning. In *ECCV*, 2020.

[18] Ekin Dogus Cubuk, Barret Zoph, Jon Shlens, and Quoc Le. Randaugment: Practical automated data augmentation with a reduced search space. In H. Larochelle, M. Ranzato, R. Hadsell, M. F. Balcan, and H. Lin, editors, *Advances in Neural Information Processing Systems*, volume 33, pages 18613–18624. Curran Associates, Inc., 2020.

[19] Chaorui Deng, Shizhe Chen, Da Chen, Yuan He, and Qi Wu. Sketch, ground, and refine: Top-down dense video captioning. In *Proceedings of the IEEE/CVF Conference on Computer Vision and Pattern Recognition*, pages 234–243, 2021.

[20] Alexey Dosovitskiy, Lucas Beyer, Alexander Kolesnikov, Dirk Weissenborn, Xiaohua Zhai, Thomas Unterthiner, Mostafa Dehghani, Matthias Minderer, Georg Heigold, Sylvain Gelly, Jakob Uszkoreit, and Neil Houlsby. An image is worth 16x16 words: Transformers for image recognition at scale. *ICLR*, 2021.

[21] Haoqi Fan, Bo Xiong, Karttikeya Mangalam, Yanghao Li, Zhicheng Yan, Jitendra Malik, and Christoph Feichtenhofer. Multiscale vision transformers. In *ICCV*, 2021.

[22] C. Feichtenhofer, H. Fan, J. Malik, and K. He. Slowfast networks for video recognition. In *2019 IEEE/CVF International Conference on Computer Vision (ICCV)*, pages 6201–6210, 2019.

[23] Chen Gao, Jiarui Xu, Yuliang Zou, and Jia-Bin Huang. Drg: Dual relation graph for human-object interaction detection. *ArXiv*, abs/2008.11714, 2020.

[24] Lianli Gao, Zhao Guo, Hanwang Zhang, Xing Xu, and Heng Tao Shen. Video captioning with attention-based lstm and semantic consistency. *IEEE Transactions on Multimedia*, 19(9):2045–2055, 2017.

[25] Rohit Girdhar, Joao Carreira, Carl Doersch, and Andrew Zisserman. Video action transformer network. In *Proceedings of the IEEE Conference on Computer Vision and Pattern Recognition*, pages 244–253, 2019.

[26] Rohit Girdhar, Deva Ramanan, Abhinav Kumar Gupta, Josef Sivic, and Bryan C. Russell. Actionvlad: Learning spatio-temporal aggregation for action classification. *2017 IEEE Conference on Computer Vision and Pattern Recognition (CVPR)*, pages 3165–3174, 2017.

[27] Rohit Girdhar, Mannat Singh, Nikhila Ravi, Laurens van der Maaten, Armand Joulin, and Ishan Misra. Omnivore: A Single Model for Many Visual Modalities. In *CVPR*, 2022.

[28] Chengyue Gong, Dilin Wang, Meng Li, Vikas Chandra, and Qiang Liu. Vision transformers with patch diversification, 2021.

[29] Priya Goyal, Dhruv Kumar Mahajan, Abhinav Kumar Gupta, and Ishan Misra. Scaling and benchmarking self-supervised visual representation learning. *2019 IEEE/CVF International Conference on Computer Vision (ICCV)*, pages 6390–6399, 2019.

[30] Raghav Goyal, Samira Ebrahimi Kahou, Vincent Michalski, Joanna Materzynska, Susanne Westphal, Heuna Kim, Valentin Haenel, Ingo Fruend, Peter Yianilos, Moritz Mueller-Freitag, et al. The" something something" video database for learning and evaluating visual common sense. In *ICCV*, page 5, 2017.

[31] Kristen Grauman, Andrew Westbury, Eugene Byrne, Zachary Chavis, Antonino Furnari, Rohit Girdhar, Jackson Hamburger, Hao Jiang, Miao Liu, Xingyu Liu, Miguel Martin, Tushar Nagarajan, Ilija Radosavovic, Santhosh Kumar Ramakrishnan, Fiona Ryan, Jayant Sharma, Michael Wray, Mengmeng Xu, Eric Zhongcong Xu, Chen Zhao, Siddhant Bansal, Dhruv Batra, Vincent Cartillier, Sean Crane, Tien Do, Morrie Doulaty, Akshay Erapalli, Christoph Feichtenhofer, Adriano Fragomeni, Qichen Fu, Christian Fuegen, Abrham Gebreselasie, Cristina Gonzalez, James Hillis, Xuhua Huang, Yifei Huang, Wenqi Jia, Weslie Khoo, Jachym Kolar, Satwik Kottur, Anurag Kumar, Federico Landini, Chao Li, Yanghao Li, Zhenqiang Li, Karttikeya Mangalam, Raghava Modhugu, Jonathan Munro, Tullie Murrell, Takumi Nishiyasu, Will Price, Paola Ruiz Puentes, Merey Ramazanova, Leda Sari, Kiran Somasundaram, Audrey Southerland, Yusuke Sugano, Ruijie Tao, Minh Vo, Yuchen Wang, Xindi Wu, Takuma Yagi, Yunyi Zhu, Pablo Arbelaez, David Crandall, Dima Damen, Giovanni Maria Farinella, Bernard Ghanem, Vamsi Krishna Ithapu, C. V. Jawahar, Hanbyul Joo, Kris Kitani, Haizhou Li, Richard Newcombe, Aude Oliva,

Hyun Soo Park, James M. Rehg, Yoichi Sato, Jianbo Shi, Mike Zheng Shou, Antonio Torralba, Lorenzo Torresani, Mingfei Yan, and Jitendra Malik. Ego4d: Around the World in 3,000 Hours of Egocentric Video. *CoRR*, abs/2110.07058, 2021.

[32] Chunhui Gu, Chen Sun, David A. Ross, Carl Vondrick, Caroline Pantofaru, Yeqing Li, Sudheendra Vijayanarasimhan, George Toderici, Susanna Ricco, Rahul Sukthankar, Cordelia Schmid, and Jitendra Malik. AVA: A video dataset of spatio-temporally localized atomic visual actions. In *2018 IEEE Conference on Computer Vision and Pattern Recognition, CVPR 2018, Salt Lake City, UT, USA, June 18-22, 2018*, pages 6047–6056. IEEE Computer Society, 2018.

[33] Roei Herzig, Amir Bar, Huijuan Xu, Gal Chechik, Trevor Darrell, and Amir Globerson. Learning canonical representations for scene graph to image generation. In *European Conference on Computer Vision*, 2020.

[34] Roei Herzig, Elad Ben-Avraham, Karttikeya Mangalam, Amir Bar, Gal Chechik, Anna Rohrbach, Trevor Darrell, and Amir Globerson. Object-region video transformers. In *Conference on Computer Vision and Pattern Recognition (CVPR)*, 2022.

[35] Roei Herzig, Elad Levi, Huijuan Xu, Hang Gao, Eli Brosh, Xiaolong Wang, Amir Globerson, and Trevor Darrell. Spatio-temporal action graph networks. In *Proceedings of the IEEE International Conference on Computer Vision Workshops*, pages 0–0, 2019.

[36] Roei Herzig, Moshiko Raboh, Gal Chechik, Jonathan Berant, and Amir Globerson. Mapping images to scene graphs with permutation-invariant structured prediction. In *Advances in Neural Information Processing Systems (NIPS)*, 2018.

[37] Geoffrey E. Hinton, Nitish Srivastava, A. Krizhevsky, Ilya Sutskever, and R. Salakhutdinov. Improving neural networks by preventing co-adaptation of feature detectors. *ArXiv*, abs/1207.0580, 2012.

[38] Max Jaderberg, Volodymyr Mnih, Wojciech M. Czarnecki, Tom Schaul, Joel Z. Leibo, David Silver, and Koray Kavukcuoglu. Reinforcement learning with unsupervised auxiliary tasks. *ArXiv*, abs/1611.05397, 2017.

[39] Andrew Jaegle, Sebastian Borgeaud, Jean-Baptiste Alayrac, Carl Doersch, Catalin Ionescu, David Ding, Skanda Koppula, Daniel Zoran, Andrew Brock, Evan Shelhamer, et al. Perceiver io: A general architecture for structured inputs & outputs. *arXiv preprint arXiv:2107.14795*, 2021.

[40] Andrew Jaegle, Felix Gimeno, Andy Brock, Oriol Vinyals, Andrew Zisserman, and Joao Carreira. Perceiver: General perception with iterative attention. In *International Conference on Machine Learning*, pages 4651–4664. PMLR, 2021.

[41] Achiya Jerbi, Roei Herzig, Jonathan Berant, Gal Chechik, and Amir Globerson. Learning object detection from captions via textual scene attributes. *ArXiv*, abs/2009.14558, 2020.

[42] Jingwei Ji, Ranjay Krishna, Li Fei-Fei, and Juan Carlos Niebles. Action genome: Actions as composition of spatio-temporal scene graphs. *arXiv preprint arXiv:1912.06992*, 2019.

[43] Justin Johnson, Agrim Gupta, and Li Fei-Fei. Image generation from scene graphs. In *Proceedings of the IEEE conference on computer vision and pattern recognition*, pages 1219–1228, 2018.

[44] Keizo Kato, Yin Li, and Abhinav Gupta. Compositional learning for human object interaction. In *ECCV*, 2018.

[45] Will Kay, Joao Carreira, Karen Simonyan, Brian Zhang, Chloe Hillier, Sudheendra Vijayanarasimhan, Fabio Viola, Tim Green, Trevor Back, Paul Natsev, et al. The kinetics human action video dataset. *arXiv preprint arXiv:1705.06950*, 2017.

[46] Diederik P Kingma and Jimmy Ba. Adam: A method for stochastic optimization. *arXiv preprint arXiv:1412.6980*, 2014.

[47] Thomas N Kipf and Max Welling. Semi-supervised classification with graph convolutional networks. *arXiv preprint arXiv:1609.02907*, 2016.

[48] Ranjay Krishna, Ines Chami, Michael S. Bernstein, and Li Fei-Fei. Referring relationships. *ECCV*, 2018.

[49] Ranjay Krishna, Yuke Zhu, Oliver Groth, Justin Johnson, Kenji Hata, Joshua Kravitz, Stephanie Chen, Yannis Kalantidis, Li-Jia Li, David A Shamma, et al. Visual genome: Connecting language and vision using crowdsourced dense image annotations. *International Journal of Computer Vision*, 123(1):32–73, 2017.

[50] Dongxu Li, Junnan Li, Hongdong Li, Juan Carlos Niebles, and Steven C. H. Hoi. Align and prompt: Video-and-language pre-training with entity prompts. *ArXiv*, abs/2112.09583, 2021.

[51] Liunian Harold Li, Mark Yatskar, Da Yin, Cho-Jui Hsieh, and Kai-Wei Chang. Visualbert: A simple and performant baseline for vision and language. *ArXiv*, abs/1908.03557, 2019.

[52] Xiujun Li, Xi Yin, Chunyuan Li, Xiaowei Hu, Pengchuan Zhang, Lei Zhang, Lijuan Wang, Houdong Hu, Li Dong, Furu Wei, Yejin Choi, and Jianfeng Gao. Oscar: Object-semantics aligned pre-training for vision-language tasks. *ECCV 2020*, 2020.

[53] Yanghao Li, Chao-Yuan Wu, Haoqi Fan, Karttikeya Mangalam, Bo Xiong, Jitendra Malik, and Christoph Feichtenhofer. Mvitv2: Improved multiscale vision transformers for classification and detection. In *CVPR*, 2022.

[54] Yingwei Li, Yi Li, and Nuno Vasconcelos. Resound: Towards action recognition without representation bias. In *Proceedings of the European Conference on Computer Vision (ECCV)*, September 2018.

[55] Junwei Liang, Lu Jiang, Juan Carlos Niebles, Alexander G. Hauptmann, and Li Fei-Fei. Peeking into the future: Predicting future person activities and locations in videos. *2019 IEEE/CVF Conference on Computer Vision and Pattern Recognition (CVPR)*, pages 5718–5727, 2019.

[56] Ze Liu, Jia Ning, Yue Cao, Yixuan Wei, Zheng Zhang, Stephen Lin, and Han Hu. Video swin transformer. *arXiv preprint arXiv:2106.13230*, 2021.

[57] Jiasen Lu, Vedanuj Goswami, Marcus Rohrbach, Devi Parikh, and Stefan Lee. 12-in-1: Multi-task vision and language representation learning. *2020 IEEE/CVF Conference on Computer Vision and Pattern Recognition (CVPR)*, pages 10434–10443, 2020.

[58] Chih-Yao Ma, Asim Kadav, Iain Melvin, Zsolt Kira, Ghassan Al-Regib, and Hans Peter Graf. Attend and interact: Higher-order object interactions for video understanding. *2018 IEEE/CVF Conference on Computer Vision and Pattern Recognition*, pages 6790–6800, 2018.

[59] Joanna Materzynska, Tete Xiao, Roei Herzig, Huijuan Xu, Xiaolong Wang, and Trevor Darrell. Something-else: Compositional action recognition with spatial-temporal interaction networks. In *proceedings of the IEEE Conference on Computer Vision and Pattern Recognition*, 2020.

[60] E. Mavroudi, Benjamín Béjar Haro, and René Vidal. Representation learning on visual-symbolic graphs for video understanding. In *ECCV*, 2020.

[61] Antoine Miech, Jean-Baptiste Alayrac, Lucas Smaira, Ivan Laptev, Josef Sivic, and Andrew Zisserman. End-to-end learning of visual representations from uncurated instructional videos. In *Proceedings of the IEEE/CVF Conference on Computer Vision and Pattern Recognition*, pages 9879–9889, 2020.

[62] Antoine Miech, Dimitri Zhukov, Jean-Baptiste Alayrac, Makarand Tapaswi, Ivan Laptev, and Josef Sivic. Howto100m: Learning a text-video embedding by watching hundred million narrated video clips. In *Proceedings of the IEEE/CVF International Conference on Computer Vision*, pages 2630–2640, 2019.

[63] Tushar Nagarajan, Yanghao Li, Christoph Feichtenhofer, and Kristen Grauman. Ego-topo: Environment affordances from egocentric video. *2020 IEEE/CVF Conference on Computer Vision and Pattern Recognition (CVPR)*, pages 160–169, 2020.

[64] Andrew Owens, Jiajun Wu, Josh H McDermott, William T Freeman, and Antonio Torralba. Learning sight from sound: Ambient sound provides supervision for visual learning. *International Journal of Computer Vision*, 126(10):1120–1137, 2018.

[65] Yingwei Pan, Ting Yao, Houqiang Li, and Tao Mei. Video captioning with transferred semantic attributes. In *Proceedings of the IEEE conference on computer vision and pattern recognition*, pages 6504–6512, 2017.

[66] Mandela Patrick, Dylan Campbell, Yuki M. Asano, Ishan Misra Florian Metze, Christoph Feichtenhofer, Andrea Vedaldi, and Joao F. Henriques. Keeping your eye on the ball: Trajectory attention in video transformers, 2021.

[67] Moshiko Raboh, Roei Herzig, Gal Chechik, Jonathan Berant, and Amir Globerson. Differentiable scene graphs. In *WACV*, 2020.

[68] Alec Radford, Jong Wook Kim, Chris Hallacy, Aditya Ramesh, Gabriel Goh, Sandhini Agarwal, Girish Sastry, Amanda Askell, Pamela Mishkin, Jack Clark, et al. Learning transferable visual models from natural language supervision. In *International Conference on Machine Learning*, pages 8748–8763. PMLR, 2021.

[69] Hamid Rezatofighi, Nathan Tsoi, JunYoung Gwak, Amir Sadeghian, Ian Reid, and Silvio Savarese. Generalized intersection over union. In *The IEEE Conference on Computer Vision and Pattern Recognition (CVPR)*, June 2019.

[70] Arka Sadhu, Tanmay Gupta, Mark Yatskar, Ramakant Nevatia, and Aniruddha Kembhavi. Visual semantic role labeling for video understanding. *2021 IEEE/CVF Conference on Computer Vision and Pattern Recognition (CVPR)*, pages 5585–5596, 2021.

[71] Adam Santoro, David Raposo, David G Barrett, Mateusz Malinowski, Razvan Pascanu, Peter Battaglia, and Timothy Lillicrap. A simple neural network module for relational reasoning. In I. Guyon, U. V. Luxburg, S. Bengio, H. Wallach, R. Fergus, S. Vishwanathan, and R. Garnett, editors, *Advances in Neural Information Processing Systems*, volume 30. Curran Associates, Inc., 2017.

[72] Dandan Shan, Jiaqi Geng, Michelle Shu, and David Fouhey. Understanding human hands in contact at internet scale. In *CVPR*, 2020.

[73] Trevor Scott Standley, Amir Roshan Zamir, Dawn Chen, Leonidas J. Guibas, Jitendra Malik, and Silvio Savarese. Which tasks should be learned together in multi-task learning? In *ICML*, 2020.

[74] Weijie Su, Xizhou Zhu, Yue Cao, Bin Li, Lewei Lu, Furu Wei, and Jifeng Dai. Vl-bert: Pre-training of generic visual-linguistic representations. In *International Conference on Learning Representations*, 2020.

[75] Chen Sun, Abhinav Shrivastava, Carl Vondrick, Kevin Murphy, Rahul Sukthankar, and Cordelia Schmid. Actor-centric relation network. In *Proceedings of the European Conference on Computer Vision (ECCV)*, pages 318–334, 2018.

[76] Xu Sun, Tongwei Ren, Yuan Zi, and Gangshan Wu. Video visual relation detection via multimodal feature fusion. *Proceedings of the 27th ACM International Conference on Multimedia*, 2019.

[77] Hao Tan and Mohit Bansal. Lxmert: Learning cross-modality encoder representations from transformers. In *EMNLP/IJCNLP (1)*, pages 5099–5110. Association for Computational Linguistics, 2019.

[78] Makarand Tapaswi, Yukun Zhu, Rainer Stiefelhagen, Antonio Torralba, Raquel Urtasun, and Sanja Fidler. Movieqa: Understanding stories in movies through question-answering. In *Proceedings of the IEEE conference on computer vision and pattern recognition*, pages 4631–4640, 2016.

[79] Bugra Tekin, Federica Bogo, and Marc Pollefeys. H+o: Unified egocentric recognition of 3d hand-object poses and interactions. *2019 IEEE/CVF Conference on Computer Vision and Pattern Recognition (CVPR)*, pages 4506–4515, 2019.

[80] Xiaolong Wang and Abhinav Gupta. Videos as space-time region graphs. In *ECCV*, 2018.

[81] Xin Wang, Wenhu Chen, Jiawei Wu, Yuan-Fang Wang, and William Yang Wang. Video captioning via hierarchical reinforcement learning. In *Proceedings of the IEEE Conference on Computer Vision and Pattern Recognition (CVPR)*, June 2018.

[82] Chao-Yuan Wu and Philipp Krähenbühl. Towards Long-Form Video Understanding. In *CVPR*, 2021.

[83] Bingjie Xu, Yongkang Wong, Junnan Li, Qi Zhao, and M. Kankanhalli. Learning to detect human-object interactions with knowledge. *2019 IEEE/CVF Conference on Computer Vision and Pattern Recognition (CVPR)*, pages 2019–2028, 2019.

[84] Keyulu Xu, Jingling Li, Mozhi Zhang, Simon S. Du, Ken ichi Kawarabayashi, and Stefanie Jegelka. What can neural networks reason about? In *International Conference on Learning Representations*, 2020.

[85] Vinicius Zambaldi, David Raposo, Adam Santoro, Victor Bapst, Yujia Li, Igor Babuschkin, Karl Tuyls, David Reichert, Timothy Lillicrap, Edward Lockhart, et al. Relational deep reinforcement learning. *arXiv preprint arXiv:1806.01830*, 2018.

[86] Bowen Zhang, Jiahui Yu, Christopher Fifty, Wei Han, Andrew M. Dai, Ruoming Pang, and Fei Sha. Co-training transformer with videos and images improves action recognition. *ArXiv*, abs/2112.07175, 2021.

[87] Chuhan Zhang, Ankush Gputa, and Andrew Zisserman. Temporal query networks for fine-grained video understanding. In *Conference on Computer Vision and Pattern Recognition (CVPR)*, 2021.

[88] Yubo Zhang, Pavel Tokmakov, Martial Hebert, and Cordelia Schmid. A structured model for action detection. In *2019 IEEE/CVF Conference on Computer Vision and Pattern Recognition (CVPR)*, pages 9967–9976, 2019.

# Supplementary Material for "Bringing Image Scene Structure to Video via Frame-Clip Consistency of Object Tokens"

In this supplementary file, we provide additional information about our experimental results, qualitative examples, implementation details, and datasets. Specifically, Section A provides more experiment results, Section B provides qualitative visualizations to illustrate our approach, Section C provides additional implementation details, and Section D provides additional datasets details.

## A  Additional Experiment Results

First, we discuss the pretraining and finetuning of the SViT model variants (Section A.1). Next, we present additional ablations (Section A.2) we performed in order to test the contribution of the different SViT components.

### A.1  SViT Variants

As explained in Section 3.3, we use several SViT variants, see Table 5 for details. Additionally, Table 7 provides a detailed listing of the datasets we used for each SViT variant and task.

| Variant | Pretrain Auxiliary Image Data | Finetune Auxiliary Image Data |
|---|---|---|
| SViT-ID | In-Domain + Different-Domain | In-Domain |
| SViT-DD | Different-Domain | Different-Domain |
| SViT-SFT | Different-Domain | None |

Table 5: **SViT Variants**

### A.2  Additional Experiments

Next, we provide additional experiments in Table 6.

**Pretraining and finetuning using in-domain auxiliary images**. As part of finetuning, SViT-ID makes use of in-domain auxiliary image data, whereas for pretraining, it uses a different domain and in-domain of image data (see Table 5). It is interesting to see what happens if one uses only in-domain data for both pretraining and finetuning, and no data from external datasets. Towards this end, we consider a SViT training setup where only the in-domain images are used for both pre-training and fine-tuning. We do this both for SmthElse and Ego4D. Results are shown in Tables 6a and 6b. It can be seen that results with only in-domain image data (both pretraining and finetuning) are quite close to those of using both different and in-domain data. Thus, we conclude that the performance improvement of the in-domain and different domain are similar, demonstrating that our method is capable of benefiting both from in-domain and different-domain data.

**Pertraining and Finetuning with or without auxiliary images**. In Table 6d, we check the effect of using the auxiliary image data during the pretraining and finetuning for the standard SViT-ID model. Namely, during the pretraining and fine tuning processes, we examine each combination of using or not using auxiliary image data. We can see that doing finetuning and pretraining with auxiliary images yields the best results (65.8). Alternatively, only pretraining with auxiliary images (64.2) or only finetuning with auxiliary images (64.4) are advantageous compared to not using any auxiliary images. In fact, not using images at all is equivalent to MViT-v2 with additional tokens, and has no benefit over MViT-v2.

**Consistency Loss**. In Table 6f we consider two different types of frame-clip consistencies. The first is patch token consistency, which is simply replacing the object tokens with the patch tokens in the frame-clip consistency loss and the second is object token consistency (as described in Equation 7 in the main paper). Comparing the results to not using any frame-clip consistency (65.0), we note that object token consistency is beneficial (+0.8), but patch consistency decreases the performance (-6.3). We hypothesize that patch consistency may reduce patch diversification within the frames, thus reducing performance. A recent study [28] has demonstrated that vision transformers tend

### (a) ID pretraining on **SomethingElse**

| Model | Pretraining Aux. Images | Top-1 |
|---|---|---|
| SViT-ID | diff-domain | 65.8 |
| SViT ×2% | diff-domain | 65.1 |
| SViT-ID | in-domain | 65.6 |
| SViT ×2% | in-domain | 65.1 |

### (b) ID pretraining on **Ego4D**

| Model | Pretraining Aux. Images | Temporal Loc. Error | Cls. Top-1 |
|---|---|---|---|
| SViT-ID | diff-domain | 0.64 | 73.8 |
| SViT-ID | in-domain | 0.642 | 71.9 |

### (c) HAOG Attributes

| Model | Top-1 |
|---|---|
| SViT -Contact | 65.5 |
| SViT -Contact -Corresp. | 65.3 |
| SViT +Geometry | 65.7 |
| SViT | 65.8 |

### (d) Auxiliary images during pretraining and finetuning

| Model | Pretrain | Finetune | Top-1 |
|---|---|---|---|
| SViT-ID | w/o Images | w/o Images | 63.4 |
| SViT-ID | w/o Images | w/ Images | 64.4 |
| SViT-ID | w/ Images | w/o Images | 64.2 |
| SViT-ID | w/ Images | w/ Images | 65.8 |

### (e) Images-Videos Ratio

| Images/Videos | Top-1 |
|---|---|
| 1/1 | 65.6 |
| 1/10 | 64.8 |
| 1/100 | 61.6 |

### (f) Consistency loss

| Model | Frame-Clip Consistency Type | Top-1 |
|---|---|---|
| SViT-ID | Patches | 58.7 |
| SViT-ID | Objects | 65.8 |
| SViT-ID | None | 65.0 |

Table 6: **Additional Experiments.** Unless otherwise noted, all experiments were performed on the "SomethingElse" dataset. In tables 6a and 6b, "diff-domain" refers to different-domain.

to map different patches into similar latent representations, which results in information loss and performance degradation.

**The minimum requirements to make the model work**. In Table 6e we aim to systematically demonstrate the minimum requirements necessary for the model to function reasonably well. We explored the minimum requirements to make our model work. We provide the following experiments: (i) A ratio of 1 image to X videos (where X is 1,10, and 100): 65.6 (1-to-1), 64.8 (1-to-10), and 61.55 (1-to-100). The results indicate that using too few annotated images (1 image per 100 videos) may result in degradation, but using a relatively small number of images (1 image per 10 videos) is sufficient to achieve good improvement.

**HAOG attributes**. We use the object tokens to predict several aspects of the HAOG. In Table 6c we consider several variations on this prediction. The model "SViT-Contact" does not predict the contact information, resulting in $-0.3\%$ drop compared to SViT. The model "SViT+Geometry" adds to SViT additional heads that predict distances between all HAOG objects (to explicitly add geometric bias). This does not affect performance. Finally, we explore what happens when we do not provide the information about the identity of bounding boxes in the training data. Namely, we treat them as four boxes, and ask the model to match those to the object tokens, via a matching losss as in [12]. This is model "SViT-Contact -Corresp." which is quite close to "SViT-Contact", indicating that SViT can perhaps be trained with weaker labels.

**Frame-clip consistency loss**. In order to validate frame-clip consistency loss, we evaluate the following experiments: (i) We perturb the image temporal position without consistency loss. We refer to this version as SViT-Perturb (and similarly SViT-Perturb-DD and SViT-Perturb-ID). (ii) The HAOG annotations are extrapolated from one random frame of a video. This serves as additional supervision (without the consistency loss) since the HAOG annotations correspond to the video frames (we note that SViT does not require such correspondence since it uses only HAOG annotations from single images). (iii) We predict the HAOG of a random frame in a video, and then duplicate it over the temporal dimension and use it in the same manner as in the consistency loss. We find that these three baselines lead to worse performance. For the first experiment, we obtained SViT-Perturb-DD with 64.1 (while SViT-DD obtained 65.1), and SViT-Perturb-ID with 65.2 (while SViT-ID with consistency loss got 65.8). For the second experiment, the proposed baseline achieved 65.0 compared to our SViT-ID, which achieved 65.8 (and does not require correspondence). Taken together, this demonstrates the importance of our frame-clip consistency loss.

**HAOG "distillation"**. In this experiment we first train a vision transformer to predict the HAOG task alone and then fine-tune its backbone network for the video-related task. We evaluated and achieved 62.3 compared to 63.3 of MViTv2. This indicates that training HAOGs as a form of "distillation", as we do in SViT, is indeed important.

**The object tokens representations**. To verify what the object tokens learned, we can evaluate the ability of object tokens to be utilized explicitly for the auxiliary task as a simple detector of hands and objects in images. This is accomplished by predicting the detections on SomethingElse based on the learned object tokens. The learned object tokens were compared to the MViTv2 model extended with regression and detection heads. Our model achieved an mAP of 16.8, while the proposed baseline achieved a similar result with an mAP of 15.5. These results suggest that the object tokens learn meaningful and useful representations.

**Computational cost**. The cost computation of our approach is relatively small compared to MViTv2: +0.4% in Giga ($10^9$) FLOPs (MViTv2 with 64.5, while SViT with 64.7) and +5% in Mega ($10^6$) parameters (MViTv2 with 34.4, while SViT with 36.1).

**Hands in Diving48**. To further analyze SViT performance on Diving48 we provide an experiment in which we remove the "hand" tokens from the HAOG annotation during training, and an experiment in which we remove the "object" tokens from the HAOG annotation during training. The result for removing the "hand" tokens is a degradation of 1.7 in top-1 accuracy, and for removing the "object" tokens is 0.6. We hypothesize that the "human" boxes provide a prior for localization that helps in classifying diving actions (diving actions could be classified based on appearance and pose).

## B  Qualitative Visualizations

We present qualitative visualizations of the SViT-ID "object tokens" in videos for the following datasets: SomethingElse, Ego4D, and Diving48 in Figure 3. As explained in the main paper, two tokens correspond to the "right hand" and "left hand", whereas the other two tokens correspond to the objects they are interacting with. We can see that the "object tokens" are used to detect relevant objects, even on Diving48, where there is no human-object interaction. In Diving48, we observed a few interesting phenomena: (i) the diving persons are recognized as hands. This may be explain by the fact that skin has shared visual features with hands. (ii) The objects recognized in low scores indicate that the scene does not contain any objects. Overall, despite the issues we raised above, our model performed well on Diving48. This is evidence that our method is robust to multiple domains that do not necessarily fit with human-object graph structure (HAOG), and that it still benefits from these learned object tokens through the attention.

## C  Additional Implementation Details

Our SViT model can be used on top of the most common video transformers (MViT [21], TimeSformer [11], Mformer [66], Video Swin [56]). For our experiments, we choose the MViTv2 [53] model because it performs well empirically. These are all implemented based on the MViTv2 [53] library (available at https://github.com/facebookresearch/mvit), and we implement SViT based on this repository. Additionally, in $\mathcal{L}_{Nodes}$ (see 3) we include weights for the $L_1$, $BCE$ and $GIoU$ losses, set to 5, 1 and 2 respectively (across all datasets). Next, we elaborate on the additional implementation details for each dataset, including details about the dataset description, optimization, and training and inference.

### C.1  Ego4D

**Dataset**. Ego4D [31] is a new large-scale dataset of more than 3,670 hours of video data, capturing the daily-life scenarios of more than 900 unique individuals from nine different countries around the world. The videos contain audio, 3D meshes of the environment, eye gaze, stereo and/or synchronized videos from multiple egocentric cameras.

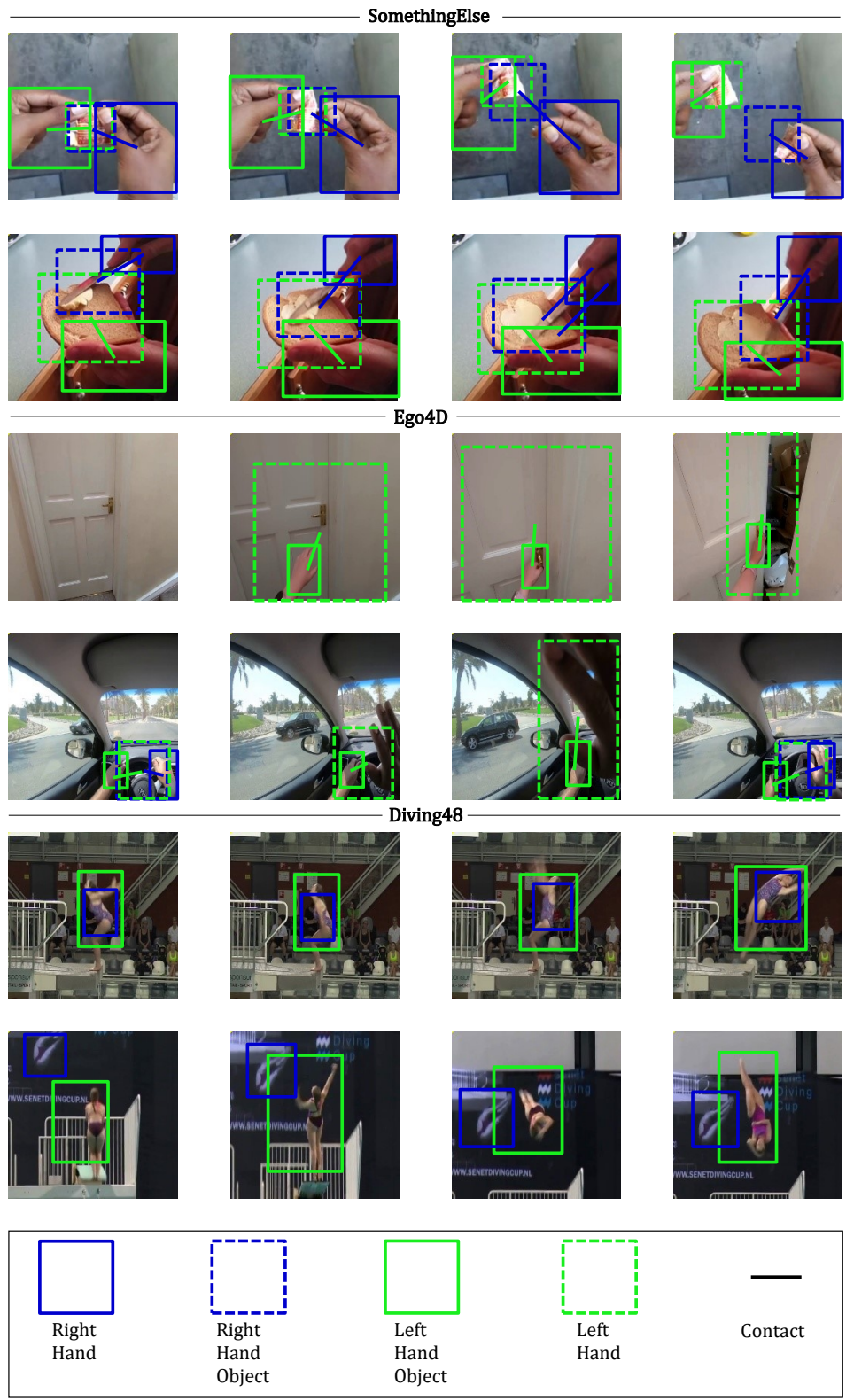

Figure 3: **Qualitative visualization of the "object tokens"**. The left object and hand prediction is visualized with **green bounding box**, the right object and hand prediction is visualized with **blue bounding box**. A line connecting between two boxes indicates a "contact" prediction between the two objects. Only boxes with a score greater than $0.6$ are plotted.

**Metrics**. In the Object State Change Temporal Localization task, the absolute error (in seconds) is used for evaluation. In the Object State Change Classification task, the top-1 accuracy is used for evaluation.

**Optimization details**. We train using 16 frames with sample rate 4 and batch-size 128 (comprising 64 videos and 64 images) on 8 RTX 3090 GPUs. We train our network for 10 epochs with Adam optimizer [46] with a momentum of $9e-1$ and Gamma $1e-1$. Following [53], we use $lr = 1e-4$ with half-period cosine decay. Additionally, we used Automatic Mixed Precision, which is implemented by PyTorch.

**Training details**. We use a standard crop size of 224, and we jitter the scales from 256 to 320. Additionally, we set $\lambda_{Con} = 10, \lambda_{HAOG} = 5, \lambda_{Vid} = 1$.

**Inference details**. We follow the official evaluation, both for the state change temporal localization and the state change classification tasks, available at https://github.com/EGO4D/hands-and-objects.

## C.2 Diving48

**Dataset**. Diving48 [54] contains 16K training and 3K testing videos spanning 48 fine-grained diving categories of diving activities. For all of these datasets, we use standard classification accuracy as our main performance metric.

**Optimization details**. We train using 16 frames with sample rate 4 and batch-size 128 (comprising 64 videos and 64 images) on 8 RTX 3090 GPUs. We train our network for 10 epochs with Adam optimizer [46] with a momentum of $9e-1$ and Gamma $1e-1$. Following [53], we use $lr = 1e-4$ with half-period cosine decay.

**Training details**. We use a standard crop size of 224 for the standard model and jitter the scales from 256 to 320. Additionally, we use RandomFlip augmentation. Finally, we sample the $T$ frames from the start and end diving annotation time, following [87]. We set $\lambda_{Con} = 10, \lambda_{HAOG} = 5, \lambda_{Vid} = 1$.

**Inference details**. We take 3 spatial crops per single clip to form predictions over a single video in testing, as in [11].

## C.3 SomethingElse

**Dataset**. The SomethingElse dataset [59] contains 174 action categories with 54,919 training and 57,876 validation samples. The proposed compositional [59] split in this dataset provides disjoint combinations of a verb (action) and noun (object) in the training and testing sets. This split defines two disjoint groups of nouns $\{\mathcal{A}, \mathcal{B}\}$ and verbs $\{1, 2\}$. Given the splits of groups, they combine the training set as $1\mathcal{A} + 2\mathcal{B}$, while the validation set is constructed by flipping the combination into $1\mathcal{B} + 2\mathcal{A}$. In this way, different combinations of verbs and nouns are divided into training or testing splits.

**Few Shot Compositional Action Recognition**. As mentioned in Section 3.4 of the main paper, we also evaluate on the few-shot compositional action recognition task in [59]. For this setting, we have 88 *base* action categories and 86 *novel* action categories. We train on the base categories (113K/12K for training/validation) and finetune on few-shot samples from the novel categories (for 5-shot, 430/50K for training/validation; for 10-shot, 860/44K for training/validation).

**Optimization details**. We train using 16 frames with sample rate 4 and batch-size 128 (comprising 64 videos and 64 images) on 8 RTX 3090 GPUs. We train our network for 100 epochs with Adam optimizer [46] with a momentum of $9e-1$ and Gamma $1e-1$. Following [53], we use $lr = 1e-4$ with half-period cosine decay. Additionally, we used Automatic Mixed Precision, which is implemented by PyTorch.

**Regularization details**. We use weight decay of $1e^{-4}$, and a dropout [37] of $5e-1$ before the final classification.

**Training details**. We use a standard crop size of 224, and we jitter the scales from 256 to 320. Additionally, we set $\lambda_{Con} = 2, \lambda_{HAOG} = 2, \lambda_{Vid} = 1$.

**Inference details**. We take 3 spatial crops per single clip to form predictions over a single video in testing.

### C.4 Something-Something v2

**Dataset**. The SSv2 [59] contains 174 action categories of common human-object interactions.

**Optimization details**. For the standard SSv2 [59] dataset, we train using 16 frames with sample rate 4 and batch-size 128 (comprising 64 videos and 64 images) on 8 RTX 3090 GPUs. We train our network for 100 epochs with Adam optimizer [46] with a momentum of $9e-1$ and Gamma $1e-1$. Following [53], we use $lr = 1e-4$ with half-period cosine decay. Additionally, we used Automatic Mixed Precision, which is implemented by PyTorch.

**Regularization details**. We use weight decay of $1e-4$, and a dropout [37] of $5e-1$ before the final classification.

**Training details**. We use a standard crop size of 224, and we jitter the scales from 256 to 320. Additionally, we use RandAugment [18] with maximum magnitude 9. We set $\lambda_{Con} = 2, \lambda_{HAOG} = 2, \lambda_{Vid} = 1$.

**Inference details**. We take 3 spatial crops per single clip to form predictions over a single video in testing as done in [53].

### C.5 AVA

**Architecture**. SlowFast [22] and MViT-v2 [53] are using a detection architecture with a RoI Align head on top of the spatio-temporal features. We followed their implementation to allow a direct comparison. Next we elaborate on the RoI Align head proposed in SlowFast [22]. First, we extract the feature maps from our SViT model by using the RoIAlign layer. Next, we take the 2D proposal at a frame into a 3D RoI by replicating it along the temporal axis, followed by a temporal global average pooling. Then, we max-pooled the RoI features and fed them to an MLP classifier for prediction.

**Optimization details**. To allow a direct comparison, we used the same configuration as in MViT-v2 [53]. We trained 16 frames with sample rate 4, depth of 16 layers and batch-size 32 on 8 RTX 3090 GPUs. We train our network for 30 epochs with an SGD optimizer. We use $lr = 0.03$ with a weight decay of $1e-8$ and a half-period cosine schedule of learning rate decaying.

**Training details**. We use a standard crop size of 224 and we jitter the scales from 256 to 320. We use the same ground-truth boxes and proposals that overlap with ground-truth boxes by $IoU > 0.9$ as in [22]. We set $\lambda_{Con} = 0.1, \lambda_{HAOG} = 0.5, \lambda_{Vid} = 1$.

**Inference details**. We perform inference on a single clip with 16 frames. For each sample, the evaluation frame is centered in frame 8. We use a crop size of 224 in test time. We take 1 spatial crop with 10 different clips sampled randomly to aggregate predictions over a single video in testing.

## D  Additional Datasets Details

Here provide detailed information about the "auxiliary image datasets" (Section D.1) as well as the licenses and privacy policies for these datasets (Section D.2).

### D.1  Auxiliary Image Datasets

In Table 7 we explicitly describe the auxiliary image datasets used in each experiment. As an example, in row 1, we describe the data used for training the SViT-SFT model for the Compositional Action Recognition (CAR) task on the SomethingElse dataset. We note that we began by pretraining on Ego4D and 100DOH.

**Image Annotations**. The collected object boxes from SSv2, 100DOH, Ego4D and AVA are purely human annotated. In SSv2, Ego4D and AVA contact relations between the object and hand are not annotated, so we assign the closest object to the hand for each hand. The contact relations for 100DOH are available in the dataset. In Ego4D, we use the image annotated for the State Change Object Detection task.

| Video Dataset | Task | Model | Pretrain Auxiliary Images | Finetune Auxiliary Images |
|---|---|---|---|---|
| SmthElse | CAR | SViT-SFT | Ego4D, 100DOH | - |
| SmthElse | CAR | SViT-DD | Ego4D, 100DOH | Ego4D, 100DOH |
| SmthElse | CAR | SViT×2% | Ego4D, 100DOH | SmthElse×2% |
| SmthElse | CAR | SViT-ID | Ego4D, 100DOH | SmthElse |
| Ego4D | SCTL | SViT-SFT | Ego4D, 100DOH | - |
| Ego4D | SCTL | SViT-DD | 100DOH | 100DOH |
| Ego4D | SCTL | SViT×2% | Ego4D, 100DOH | Ego4D×2% |
| Ego4D | SCTL | SViT-ID | Ego4D, 100DOH | Ego4D |
| Ego4D | SCC | SViT-SFT | Ego4D, 100DOH | - |
| Ego4D | SCC | SViT-DD | 100DOH | 100DOH |
| Ego4D | SCC | SViT×2% | Ego4D, 100DOH | Ego4D×2% |
| Ego4D | SCC | SViT-ID | Ego4D, 100DOH | Ego4D |
| SSv2 | AR | SViT-SFT | Ego4D, 100DOH | - |
| SSv2 | AR | SViT-DD | Ego4D, 100DOH | Ego4D, 100DOH |
| SSv2 | AR | SViT×2% | Ego4D, 100DOH | SSv2×2% |
| SSv2 | AR | SViT-ID | Ego4D, 100DOH | SSv2 |
| Diving48 | AR | SViT-SFT | Ego4D, 100DOH | - |
| Diving48 | AR | SViT-DD | Ego4D, 100DOH | Ego4D, 100DOH |
| AVA | AD | SViT-SFT | Ego4D, 100DOH | - |
| AVA | AD | SViT-DD | Ego4D, 100DOH | Ego4D, 100DOH |
| AVA | AD | SViT×2% | Ego4D, 100DOH | AVA×2% |
| AVA | AD | SViT-ID | Ego4D, 100DOH | AVA |

Table 7: **Auxiliary Image Datasets** The table describes which image datasets were you used for which SViT setup. CAR refers to Compositional Action Recognition, AR refers to Action Recognition, AD refers to Spatio-temporal Action Detection, SCTL to State Change Temporal Localization, SCC to State Change Classification

**Image/Video annotation naming conventions**. The auxiliary image data we used is originated in video frames. The main difference between video frames and a batch of images is the temporal information. Since we do not use the temporal order of the annotated video frames, and practically use them as images, we refer to them in the paper as "image annotations".

### D.2 Licenses and Privacy

The license, PII, and consent details of each dataset are in the respective papers. In addition, we wish to emphasize that the datasets we use do not contain any harmful or offensive content, as many other papers in the field also use them. Thus, we do not anticipate a specific negative impact, but, as with any Machine Learning method, we recommend to exercise caution.